\documentclass[10pt,twocolumn,letterpaper]{article}

\usepackage{cvpr}              
\usepackage{adjustbox}
\usepackage{multirow, makecell}
\usepackage{pifont}
\usepackage{bm}
\definecolor{cvprblue}{rgb}{0.21,0.49,0.74}
\usepackage[pagebackref,breaklinks,colorlinks,allcolors=cvprblue]{hyperref}

\def\paperID{43828} 
\def\confName{CVPR}
\def\confYear{2026}

\title{Revealing Artifacts via Noise Amplification: A Novel Perspective for AI-Generated Video Detection}

\author{
	Renxi Cheng\textsuperscript{1}, Jie Gui\textsuperscript{1,2,3}, Hongsong Wang\textsuperscript{4,5}\\
    $^{1}$School of Cyber Science and Engineering, Southeast University, Nanjing 210096, China\\
    $^{2}$Purple Mountain Laboratories, Nanjing 210000, China \\
	$^{3}$Engineering Research Center of Blockchain Application, Supervision And Management\\ (Southeast University), Ministry of Education, China \\
	$^{4}$School of Computer Science and Engineering, Southeast University, Nanjing 210096, China \\
	$^5$Key Laboratory of New Generation Artificial Intelligence Technology and Its Interdisciplinary \\
	Applications (Southeast University), Ministry of Education, China \\ 
	\tt\small\{renxi, guijie, hongsongwang\}@seu.edu.cn \\
}

\begin{document}
\maketitle

\begin{abstract}
With the rapid advancement of video generation models, distinguishing between AI-generated and authentic videos has emerged as a challenging endeavor. The majority of existing research endeavors concentrate on the development of detectors for identifying samples generated by generative adversarial networks. Nevertheless, the detection of AI-generated videos, particularly those produced by text-to-video models, still remains an uncharted territory. Although state-of-the-art text-to-video models can generate realistic visual content similar to real videos, they fall short of generating the details of the images and the changes in details within the videos. Inspired by this, we address AI-generated video detection from a novel perspective of bit-planes, which can effectively describe the details or noises in images or videos. To this end, we propose a simple yet effective approach called Noise Amplification. This approach first extracts noise signals based on bit-planes, then amplifies these noise signals, and finally feeds them into the discriminator networks for video fake classification. Noise amplification is comprehensively constructed by incorporating three aspects: pixel-level intensity enhancement, region-level spatial amplification, and frame-level temporal aggregation. To evaluate methods of AI-generated video detection in challenging scenarios, we also introduce a benchmark named HardGVD. Extensive experiments on both the large-scale dataset GenVidBench and HardGVD show that our simple approach significantly outperforms state-of-the-art methods.
\end{abstract}

\section{Introduction}

\label{sec:intro}
In recent years, video generation technologies, especially Sora, have witnessed remarkable advancements. The text-to-video (T2V) and image-to-video (I2V) models~\cite{modelscope,morphstudio,blattmann2023stable,xing2024dynamicrafter,chen2023seine} can generate high-quality and long videos with intricate scenes, vivid character expressions, and complex camera movements. As the visual differences between real and AI-generated videos continue to diminish, however, the diminution of visual differences between real and AI-generated videos makes them more susceptible to malicious misuse~\cite{misuse-1,misuse-2} and disinformation~\cite{earthquake}, posing serious threats to cybersecurity and social governance.

Deepfake video detection aims to identify whether a video has been manipulated or altered using artificial intelligence (AI) techniques. This technology highly depends on AI-generated media generation methods. Traditional fake media generation methods are primarily restricted to face videos, involving techniques such as entire face synthesis~\cite{zhao2017energy}, attribute modification~\cite{berthouzoz2011framework}, identity swapping~\cite{korshunova2017fast}, or face reenactment/expression swapping~\cite{tolosana2020deepfakes,juefei2022countering}. Therefore, early deepfake technologies are primarily focused on detecting fake videos that replaced a celebrity's face or other local features in a real video~\cite{guera2018deepfake}.

With the development of deep learning methods such as Variational Autoencoders (VAEs), Generative Adversarial Networks (GANs), and Transformers, generation techniques can produce more realistic and convincing fake media content. To detect videos generated by GANs, data-driven methods require the collection of a large number of real and fake videos from the targeted GAN model. If the target model is unavailable, these methods simulate the fake video generation process~\cite{zhang2019detecting}. Such deepfake detectors demonstrate poor generalization capabilities and also exhibit a lack of robustness~\cite{kaur2024deepfake}.

Recent text-to-video generation models based on diffusion models have revolutionized the field of video generation~\cite{singer2022make,zheng2024open} . By using textual prompts, these methods can generate high-fidelity videos that are rich in contextual information, thus facilitating the creation of videos for diverse and complex scenarios. As an important problem in deepfake video detection, AI-generated video detection specifically focuses on identifying videos generated by text-to-video models.
\begin{figure*}[t]
    \centering
    \includegraphics[width=0.95\linewidth]{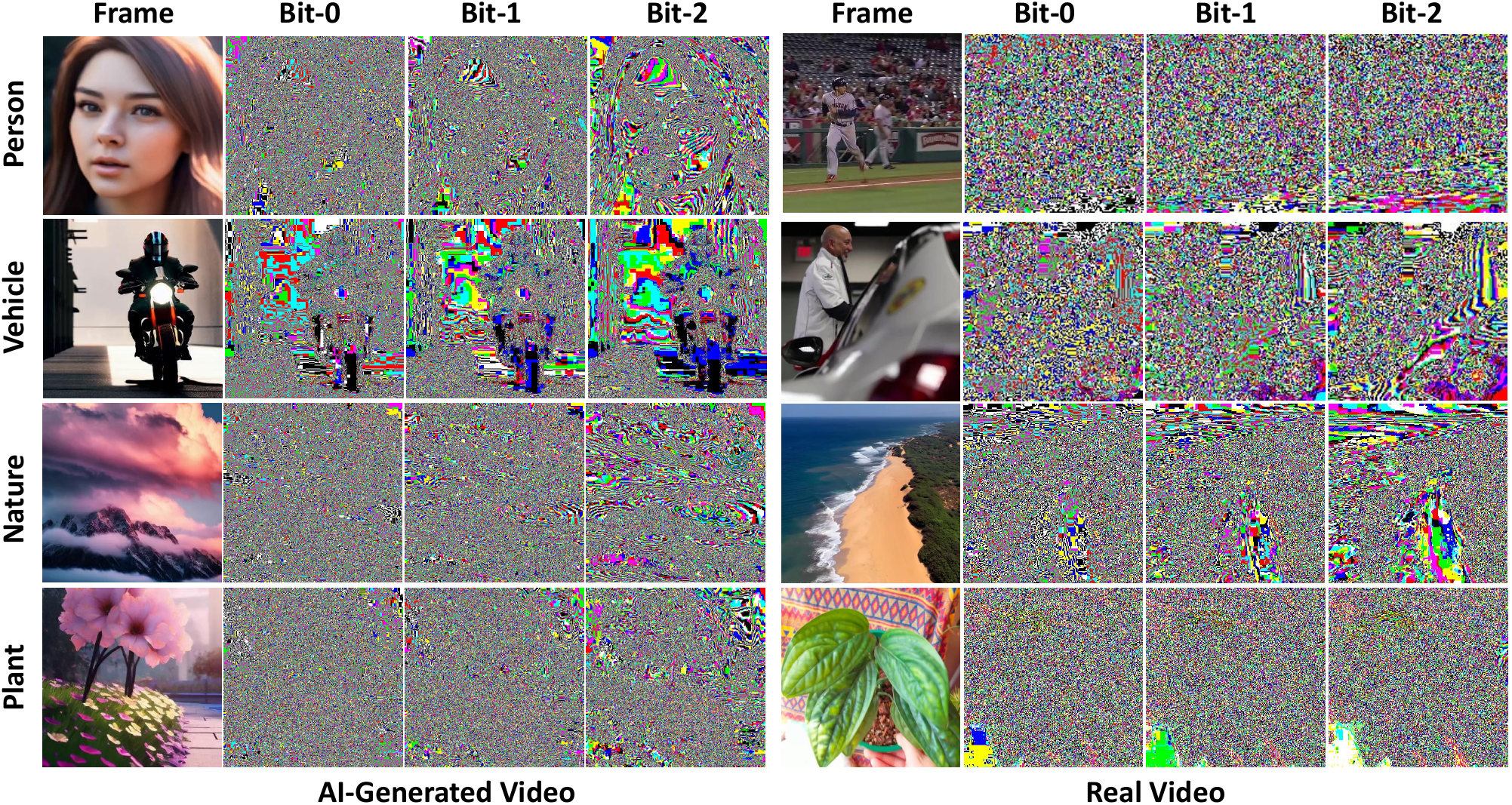}
    \caption{\textbf{Comparison of low bit-planes between real and AI-generated videos.} We exhibit the lowest three bit-planes of both real and AI-generated frames for four types of content: person, vehicle, nature, and plant. We find that artifacts in bit-planes of AI-generated frames are more clear and irregular than those in real frames. Since low bit-planes such as bit-0, bit-1 and bit-2 usually contain details, state-of-the-art generators tend to leave behind traces of clues in such low bit-planes.}
    \label{fig:intro}
\end{figure*}

Compared to AI-generated image detection~\cite{wang2023dire,luo2024lare,cozzolino2024zero}, AI-generated video detection receives much less attention in the research community. The possible reasons include the absence of standard benchmarks and the requirement for greater computational resources in AI-generated video detection. To advance the development of this field, a large-scale dataset GenVidBench~\cite{ni2025genvidbenchchallengingbenchmarkdetecting} is introduced for cross-source and cross-generator AI-generated video detection. This benchmark shows that categories such as plants, vehicles, people, buildings, and natural scenes are more likely to be misclassified by models, which can be considered hard cases. Therefore, we construct a new dataset for AI-generated video detection, named Hard Generated Video Detection (HardGVD), using prompts from these hard cases. The generators are current mainstream text-to-video generative models, including CogVideo~\cite{yang2024cogvideox}, Text2Video-Zero~\cite{text2video-zero}, Tune-A-Video~\cite{wu2023tune}, and VideoCrafter~\cite{xing2023dynamicrafter}. 

Bit-plane slicing is a technique that decomposes an image into individual bit-planes, where each plane represents a specific bit position across all pixels. Lower bit-planes reflect subtle differences in pixel values, thus having the potential to detect irregularities or artifacts that are typically absent in natural images or videos. A comparison of bit-planes between real and AI-generated video frames with various contents is illustrated in Figure~\ref{fig:intro}. Inspired by this, we address AI-generated video detection from the perspective of bit-planes.

To this end, we propose a simple pipeline, called Noise AMPlification (NAMP), for AI-generated video detection, which involves generating noise sequences from low bit-planes, amplifying the noise, and classifying them using a discriminator network. Specifically, noise amplification consists of three subsequent steps: pixel-level intensity enhancement, region-level spatial amplification, and frame-level temporal aggregation. Spatial amplification first selects an informative region in each frame to capture artifacts and then upsamples this region. Intensity enhancement directly amplifies the noise intensity, while temporal aggregation concatenates the processed noise images from different frames. Finally, standard 3D neural networks for video classification are employed as the end-to-end discriminator network, with sampled noise sequences as the input.

Our contributions can be summarized as follows: 

\begin{itemize}
    \item We study AI-generated video detection from the novel perspective of bit-plane and propose an effective method called NAMP, which amplifies details of the video through pixel intensity enhancement, spatial region amplification and temporal aggregation.
    \item We introduce the HardGVD dataset, which provides challenging cases for evaluating AI-generated video detection methods, and conduct evaluations of representative methods on this benchmark.
    \item Without any bells and whistles, our approach using I3D as the discriminator surpasses existing methods on the GenVidBench benchmark by over 10\%.
\end{itemize}

\section{Related Work}

\label{sec:related}
\label{gen_inst}

\subsection{AI-Generated Video Detection} 
Early studies focus on static spatial features within a single frame, employing traditional image processing techniques~\cite{li2018exposing}, convolutional neural network based methods~\cite{rossler2019faceforensics++} and frequency-domain based methods~\cite{frank2020leveraging}. These methods have difficulties in coping with AI-generated videos as they neglect temporal dynamics. Following works tend to seek modeling methods based on spatiotemporal information. Sabir \textit{et al.}~\cite{sabir2019recurrent} combine recurrent convolutional models and facial preprocessing techniques and achieve a significant accuracy improvement. FAST~\cite{yu2021frequency} employs the spatiotemporal Transformer to detect spatial and temporal connections combined with information from the frequency domain. The video Transformer ISTVT~\cite{zhao2023istvt} takes both spatial and temporal information into account for better generalization. 
AI-generated video detection faces new challenges due to the rapid development of video generation models, including text-to-video (T2V) and image-to-video (I2V) systems~\cite{modelscope,morphstudio,blattmann2023stable,xing2024dynamicrafter,chen2023seine} makes low-cost creation a daily routine. Modern latent diffusion models (LDMs)~\cite{Cog,modelscope,StepVideo,vc2,Animatediff,OpenSora,LTX} and recent released generative models such as WANX~\cite{wan2.1}, Hunyuan~\cite{HYV}, and Sora~\cite{Sora} further provide video creation with lower thresholds and higher realism. To cope with these challenges, Choi \textit{et al.}~\cite{choi2024exploiting} analyze the anomalous changes of style latent vectors in the time dimension to solve the problem of insufficient generalization of traditional methods as they depend on low-order spatiotemporal features. GRACE~\cite{hsu2024grace} uses spatiotemporal feature entanglement to effectively capture spatial and temporal information. Yan \textit{et al.}~\cite{yan2024generalizing} combine video-level blending data and a lightweight spatiotemporal adapter to enhance generalization and balance spatiotemporal feature learning. 

Shao \textit{et al.}~\cite{shao2023detecting} propose a novel multi-modal manipulation reasoning Transformer to comprehensively capture the fine-grained interactions between images and text. 
Bohacek \textit{et al.}~\cite{bohacek2024lost} compare information translated from lip movements and audio and take the mismatch as evidence of AI-generated content. 
Zhang \textit{et al.}~\cite{zhangphysics} introduce a novel statistic which quantifies deviations from natural video dynamics via spatial gradients and temporal density changes.
Intern{`o} \textit{et al.}~\cite{interno2025ai} hypothesize that trajectories of real‑world videos tend to be straight in neural representation space, and deviations from this straightness serve as detection cues.
Corvi \textit{et al.}~\cite{corvi2025seeing} focus on enhancing generalization based on intrinsic low‑level forensic artifacts. 
Zheng \textit{et al.}~\cite{zheng2025d3} propose a training-free method that leverages second-order temporal discrepancies between real and AI-generated videos.
Unlike these approaches, we investigate deep learning-based AI-generated image detection from the perspective of bit-planes.

\begin{figure*}[t]
    \centering
    \includegraphics[width=0.8\linewidth]{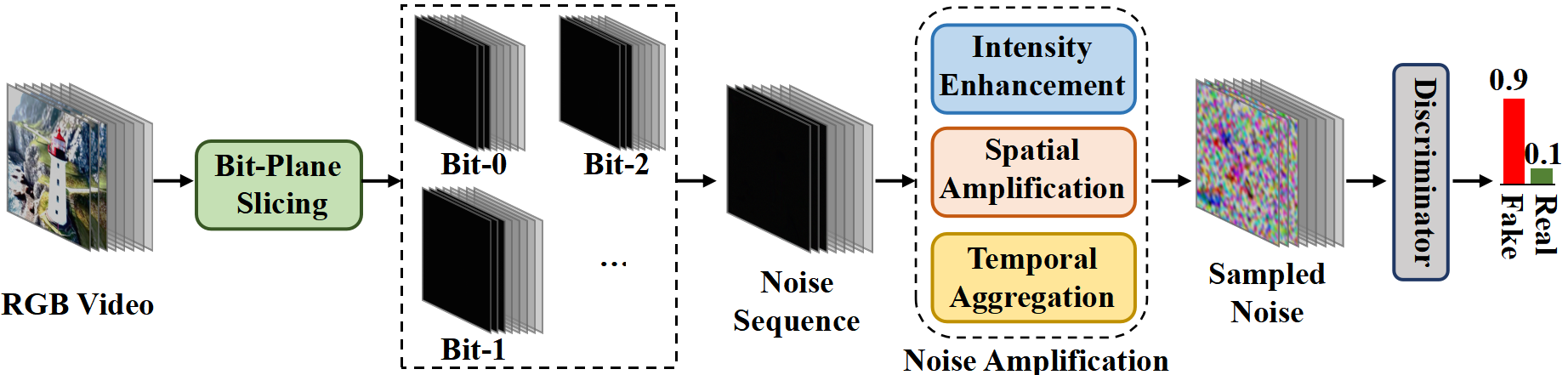}
    \caption{\textbf{Overview of the proposed pipeline.} First, frames of an RGB video are processed by bit-plane slicing to get 8 bit-planes, and low bit-plane frames (Bit-0, Bit-1, and Bit-2) are composed to produce noise sequence. Then, noise amplification of intensity enhancement, spatial amplification and temporal aggregation are applied to generate sampled noise frames. Finally, a discriminator is used to distinguish real and AI-generated videos.}
    \label{fig:method}
\end{figure*}

\subsection{AI-Generated Video Detection Dataset} 
Most existing deepfake video datasets contain only a small number of videos and are mainly used to validate algorithm feasibility. UADFV~\cite{yang2019exposing} and Deepfake TIMIT~\cite{korshunov2018deepfakes} exhibit limited generalization capabilities due to monotonous generation methods and small data scales with a total of fewer than 1000 videos. Face Forensics++~\cite{rossler2019faceforensics++} expands the generation methods to DeepFakes, Face2Face, FaceSwap and NeuralTextures. Improved datasets are introduced to enable the transition from laboratory accuracy to practical robustness. DFDC~\cite{dolhansky2020deepfake} is the first public dataset surpassing 100000 samples which simulate real scenes through volunteer selfies and diverse post-processing. Celeb-DF~\cite{li2020celeb} uses an improved GAN to generate high-fidelity frames for celebrity videos, solving the facial shaking problem. Wild-Deepfake~\cite{zi2020wilddeepfake} introduces complex conditions such as low lighting and occlusion enhancing spatiotemporal diversity. With advancements in GANs~\cite{goodfellow2014generative} and diffusion models~\cite{rombach2022high}, more datasets are based on purely generated videos. GVD~\cite{bai2024ai} contains videos generated by 11 state-of-the-art models. GenVideo~\cite{demamba} provides videos from 10 generated video models for training and videos from 10 other generated models for testing. 
GenVidBench~\cite{ni2025genvidbenchchallengingbenchmarkdetecting} is recently introduced as a large-scale dataset containing 108,430 AI-generated videos and 33,984 real videos, with all subsets generated by different generators and being heterologous via identical prompts to eliminate content impacts.

Due to the rapid development of text-to-video generation technologies, many deepfake video datasets have become outdated, and there is a lack of up-to-date AI-generated video datasets. This work introduces the HardGVD dataset, specifically designed for AI-generated video detection in challenging scenarios.

\section{Method}
\label{sec:methods}
We present a straightforward pipeline called NAMP for AI-generated video detection, as shown in Figure~\ref{fig:method}. The process begins with noise sequence extraction based on bit-planes, followed by noise amplification and end-to-end fake classification using a discriminator network.
The discriminator adopts a standard neural network architecture commonly used for video classification.

\subsection{Bit-Plane-Based Noise Extraction}
AI-generated images tend to expose unusual and unnatural artifacts in noise areas or transition regions~\cite{yan2024sanity,lota2025}. The recent AI-generated image detection approach LOTA~\cite{lota2025} demonstrates that low-bit planes contain generative artifacts that can be leveraged for detecting AI-generated content. Inspired by this, we utilize lower-order bit planes to exploit the invisible artifacts in AI-generated videos.

An RGB image can be decomposed into three grayscale images, each representing a channel $c \in \left\{ {R,G,B} \right\}$. For each gray-scale image $\bm{x^c}$, each pixel can be represented by eight weighted bits:
\begin{equation}
  \bm{x^{c}} = \sum\limits_{k = 0}^7 {2^k} \cdot \bm{x_k^{c}},
  \label{decomposition}
\end{equation}
where $\bm{x_k^{c}}$ denotes the $k$-th bit-plane of the image for the channel $c$, and $0 \leq k \leq 7$. This technique is well-known as bit plane slicing in the field of image processing.

We select the three lowest-order bit-planes, i.e., $\bm{x_2^{c}}$, $\bm{x_1^{c}}$, and $\bm{x_0^{c}}$, to simulate noise or transitional areas in an image. The low-bit image $\bm{z^{c}}$ is constructed as:
\begin{equation}
  \bm{z^{c}} = {2^2} \cdot \bm{x_2^{c}} + {2} \cdot \bm{x_1^{c}} + \bm{x_0^{c}}.
  \label{composition}
\end{equation}

\begin{figure}[htbp]
    \centering
    \includegraphics[width=\linewidth]{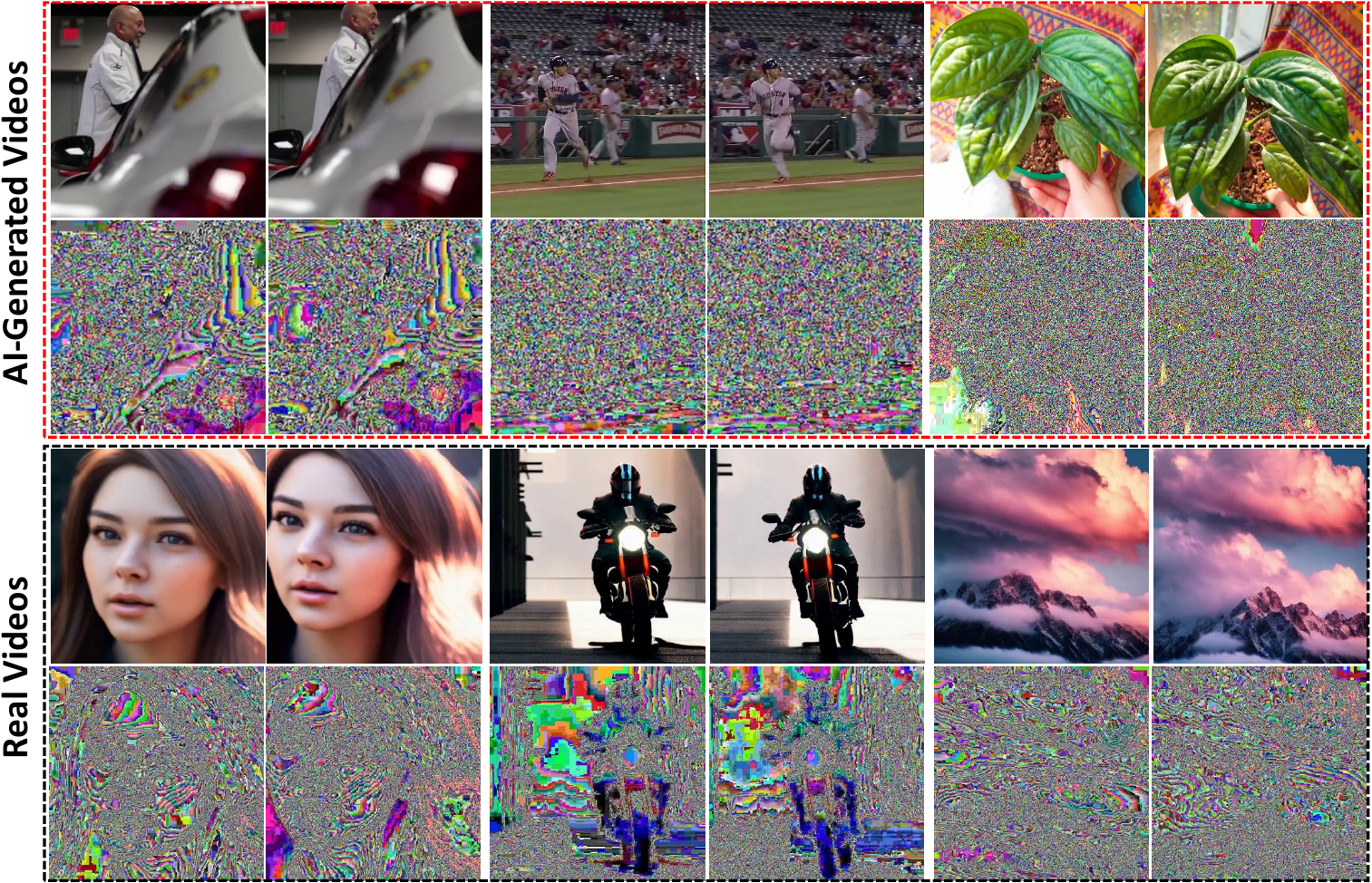}
    \caption{\textbf{Visualization of noise sequence.} We compare frames of the noise sequence between real and AI-generated videos and find that there exists more unusual artifacts in low-bit frames of AI-generated videos than those of real videos.}
    \label{fig:noise}
\end{figure}

Since a video can be decomposed into a sequence of RGB frames, the low-bit image of each frame can be similarly extracted to construct the corresponding low-bit sequence. The extracted noise sequence of both real and AI-generated videos are shown in Figure~\ref{fig:noise}. It can be seen that although these generated videos appear realistic in content, they exhibit more artifacts in the noise sequence. Therefore, we can perform AI-generated video detection based on these extracted noises.

\subsection{Noise Amplification} \label{sec:patch_select}
Since the low-bit signals or noise in videos are weak and exhibit only minor intensity variations, it is necessary to amplify and enhance these signals to facilitate neural network training. We enhance this signal from three aspects: pixel-level, region-level, and frame-level.

\noindent\textbf{Intensity Enhancement:} 
Since we only retain the lower bits of the eight bits for each pixel, the value of the noise is relatively small. For example, when using the lower three bits, the value ranges from 0 to 7. Intensity enhancement directly amplifies these weak noise signals for each pixel, scaling their values to the range of 0 to 255. 

\noindent\textbf{Spatial Amplification:} 
AI-generated videos expose unusual artifacts in some regions instead of the whole frame. These artifacts are small in the image and not easily learned to yield effective representations by neural networks. Directly resizing the whole frame to high resolution in the video leads to high computational complexity. An alternative approach is to up-sample only the saliency region of the artifacts. To achieve this goal, spatial amplification employs a divergence-based protocol to select representative regions of artifacts. To exploit artifacts with intensity changes, we randomly crop the low-bit frame $\bm{z}$ into non-overlapping patches $\bm{z_p}$ and compute the divergence-based score $g_p$ for each patch:
\begin{equation}
g_p = \sum_{i=1}^{4} ||\bm{\tilde{z}_p} * \bm{g}_i ||_1,
\label{diversity}
\end{equation}
where signs $*$ and $\|\cdot\|_1$ denote the image convolution operation and the norm of the matrix, respectively, and $\bm{g}_1, \bm{g}_2, \bm{g}_{3}$ and $\bm{g}_{4}$ are convolution kernels described as:
\begin{equation}
\begin{aligned}
\bm{g_1} &= \begin{bmatrix} -1 & 1 \end{bmatrix}, \quad\, 
\bm{g_2} = \bm{g_1}^T, \\
\quad\, \bm{g_3} &= \begin{bmatrix} -1 & 0 \\ 0 & 1 \end{bmatrix}, \quad
\bm{g_4} = \begin{bmatrix} 0 & -1 \\ 1 & 0 \end{bmatrix}.
\end{aligned}
\end{equation}

The score $g_p$ fully describes the horizontal, vertical, and diagonal diversity of a patch. To identify the patch most likely to contain artifacts, we select the patch with the maximum $g_p$. The selected small patch is then resized to the original image size. Although both our spatial amplification and ESSP~\cite{chen2024single} select a simple patch, our approach differs from ESSP in many aspects. First, our approach computes a gradient-based score instead of a texture diversity score and achieves higher efficiency through image convolution operations. Second, we select patches with the highest scores rather than the lowest ones. Third, our approach is designed for AI-generated video detection rather than AI-generated image detection.

\begin{table*}[htbp]
  
  \centering
    \begin{tabular}{l|cc|cccc|ccc}
    \toprule
    \textbf{Subset} & \textbf{Label} & \textbf{Source} & \textbf{Frames} & \textbf{FPS} & \textbf{\#Videos} & \textbf{Resolution} & \textbf{FID ↓} & \textbf{LPIPS ↓} & \textbf{VMAF ↑} \\
   \midrule
    COG & Fake & CogVideo~\cite{yang2024cogvideox} & 48 & 12 & 500 & 224$\times$224 & 22.91 & 0.14 & 83.44 \\
    T2VZ & Fake & Text2Video-Zero~\cite{text2video-zero} & 48 & 12 & 500 & 224$\times$224 & 12.24 & 0.11	& 92.13\\
    TAV & Fake & Tune-A-Video~\cite{wu2023tune} & 32 & 8 & 500 & 224$\times$224 & 24.63	& 0.13 & 91.02\\
    VC & Fake & VideoCrafter~\cite{xing2023dynamicrafter} & 48 & 12 & 500 & 256$\times$256 & 9.17 & 0.09 & 94.66\\
    YT-BI & Real & Youtube\&Bilibili & Random & Random & 2000 & Random & -- & -- & --\\
   \bottomrule
    \end{tabular}
    \caption{\textbf{The constructed HardGVD dataset.} This dataset is composed of four AI-generated subsets, COG, T2VZ, TAV, and VC, and one real subset named YT-BI. Video quality	metrics including the FID, LPIPS and VMAF are also evaluated for these subsets.} 
  \label{dataset}
\end{table*}

\begin{figure*}[htbp]
    \centering
    \includegraphics[width=\linewidth]{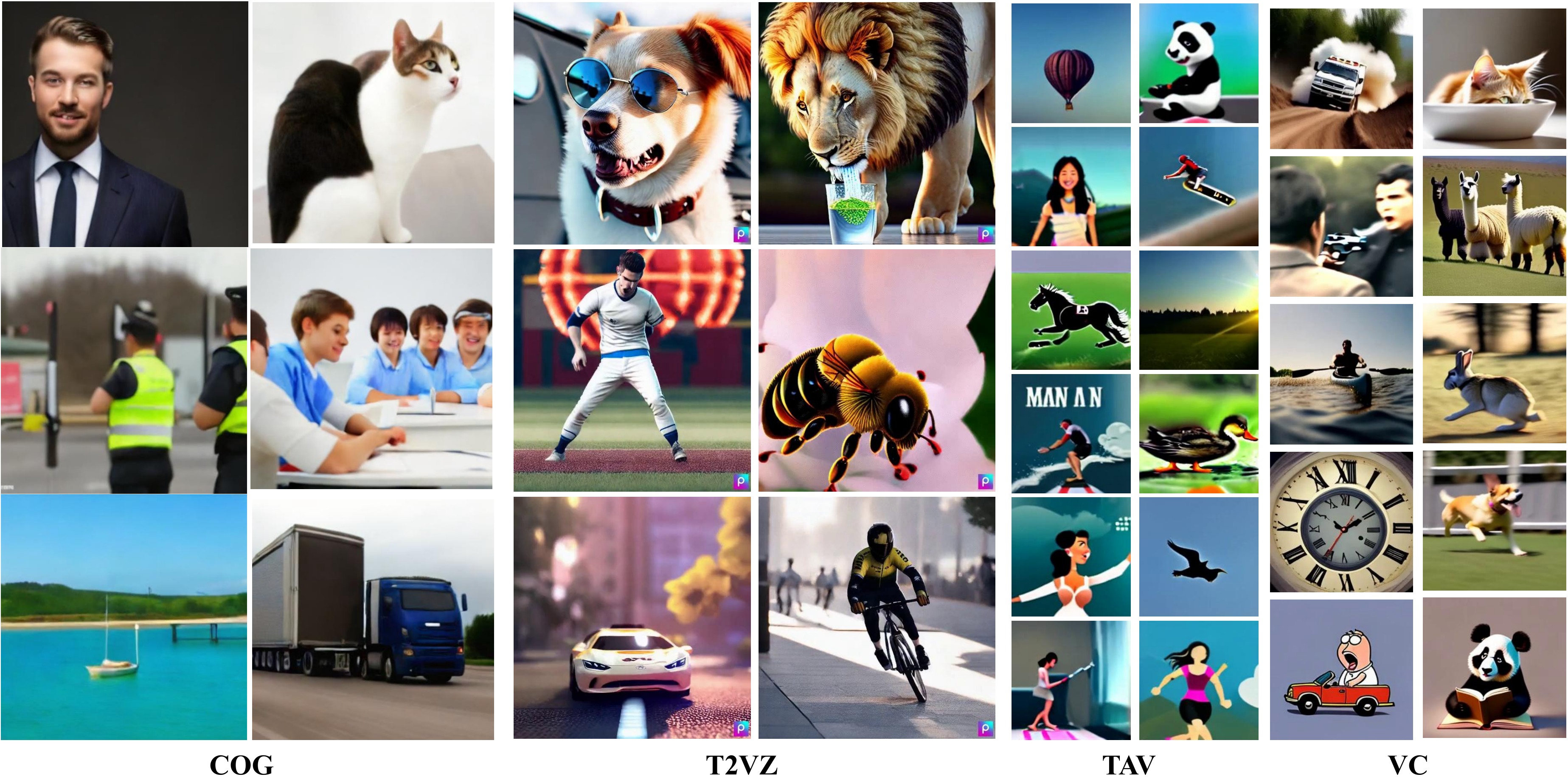}
    \caption{\textbf{Exemplar frames of the constructed HardGVD dataset.} Besides the realistic subset, this dataset contains four AI-generated subsets: COG, T2VZ, TAV, and VC, which are generated by CogVideo~\cite{yang2024cogvideox}, Text2Video-Zero~\cite{text2video-zero}, Tune-A-Video~\cite{wu2023tune}, and VideoCrafter~\cite{xing2023dynamicrafter}, respectively. Each subset contains videos of different resolutions and diverse content under challenging scenarios.}
      \label{fig:dataset}
\end{figure*}

\noindent\textbf{Temporal Aggregation:}
Since video contains rich temporal dynamics information, it is crucial to leverage this information for AI-generated video detection. For simplicity and efficiency, we directly apply image-based patch selection approach to video data. We present two simple sampling methods for video: voxel-based and frame-wise. 

The voxel-based method reuses the computed patch from the first frame across all subsequent frames, enabling the sampled sequence to retain the inherent temporal dynamics of the original video. In contrast, the frame-wise method independently performs patch selection for each frame and concatenates the resulting patches into a sequence.

\section{Experiment}

\label{sec:exp}

\subsection{Dataset}
\noindent\textbf{GenVidBench: } The GenVidBench dataset~\cite{ni2025genvidbenchchallengingbenchmarkdetecting} is a recently released large-scale AI-generated video detection benchmark with a scale of 100,000 semantic labels along with prompts and images used to generate videos. The training dataset consists of a real subset Vript~\cite{vript} and four AI-generated subsets: Pika~\cite{pika}, VideoCrafter2~\cite{vc2}, Modelscope~\cite{modelscope} and T2V-Zero~\cite{t2vz}. Videos from different substs are generated from the same set of text prompts. The testing dataset includes a real subset HD-VG~\cite{hd_vg_130m} and four AI-generated subsets: MuseV~\cite{musev}, SVD~\cite{svd}, Mora~\cite{mora} and CogVideo~\cite{cogvideo}, generated from the same set of text or image prompts.

\noindent \textbf{HardGVD: } To create more difficult cases using prompts that current models fail to classify correctly, we construct the Hard Generated Video Detection (HardGVD) dataset, which is comprised of a large number of AI-generated videos generated by several mainstream generators.

We categorize the prompts coming from GenVidBench~\cite{ni2025genvidbenchchallengingbenchmarkdetecting} into several classes, such as people, animals, natures, plants, food. According to GenVidBench~\cite{ni2025genvidbenchchallengingbenchmarkdetecting}, classes like plants, vehicles, people, buildings and natures are more likely to be misclassified by models, so we consider these classes as hard cases. Then, we use these classes to search for real videos in Youtube and Bilibili platforms, and use prompts belonging to these classes to generate AI-generated videos. We select current mainstream methods (CogVideo~\cite{yang2024cogvideox}, Text2Video-Zero~\cite{text2video-zero}, Tune-A-Video~\cite{wu2023tune} and VideoCrafter~\cite{xing2023dynamicrafter}) to generate 2000 AI-generated videos (500 for each subset), by using the same prompts in hard cases. For real videos, we crawl 500 videos from Youtube and Bilibili platforms, and select 1500 videos from HD-VG dataset~\cite{hd_vg_130m}, all of which contain classes in hard cases. To prevent bias toward majority class, we ensure that the number of fake and real videos in the entire dataset is equal. In addition to evaluating the accuracy on each subset, the averaged accuracy across the entire dataset is also assessed, as it is considered more reliable. 

Detailed statistics of the HardGVD dataset are summarized in Table~\ref{dataset}. Video quality metrics, including Fréchet Inception Distance (FID), Learned Perceptual Image Patch Similarity (LPIPS) and Video Multi-Method Assessment Fusion (VMAF) are computed for these subsets. We find that most of our subsets achieve FID scores below 25, LPIPS below 0.15, and VMAF above 90 (except for the COG). These results indicate that our HardGVD dataset demonstrates high quality in both detail preservation and temporal consistency, and closely resembles real videos. Example frames of different subsets are shown in Figure~\ref{fig:dataset}.

\subsection{Implementation Details}
We sample 16 frames of each video, and then corresponding low-bit frames are extracted. We select the patches of size 16$\times$16, and resize them to 224$\times$224. 
The learning rate is set to 0.0001, the batch size to 8, and the maximum number of training epochs to 10, with the Adam optimizer employed.

We train different models using the training set of GenVidBench~\cite{ni2025genvidbenchchallengingbenchmarkdetecting}, which includes videos generated by VideoCrafter2~\cite{vc2}, Pika~\cite{pika}, ModelScope~\cite{modelscope}, T2V-Zero~\cite{t2vz}, and Vript~\cite{vript}. Evaluation is conducted on the testing set of GenVidBench, containing videos from MuseV~\cite{musev}, SVD~\cite{svd}, Mora~\cite{mora}, CogVideo~\cite{cogvideo}, and HD-VG~\cite{hd_vg_130m}, as well as on our HardGVD dataset, which includes COG~\cite{yang2024cogvideox}, T2VZ~\cite{text2video-zero}, TAV~\cite{wu2023tune}, VC~\cite{xing2023dynamicrafter}, and YT-BI. These models are also tested on the HardGVD.

\subsection{Evaluation of AI-Generated Video Detection}

\begin{table}[tbp]
  \centering
  \resizebox{0.5\textwidth}{!}{
    \begin{tabular}{l|c|c|c|c|c|c}
    \toprule
    \textbf{Method} & \textbf{MuseV} & \textbf{SVD} & \textbf{Mora} & \textbf{CogVideo} & \textbf{HD-VG} & \textbf{Avg.} \\
   \midrule
    I3D~\cite{i3d}   & 8.15  & 8.29  & 59.24  & 60.11  & 93.99  & 49.23 \\
    SlowFast~\cite{slowfast}   & 12.25  & 12.68  & 45.93  & 38.34  & 93.63  & 41.66 \\
    F3Net~\cite{f3net}   & 37.43  & 37.27  & 39.59  & 36.46  & 52.76  & 42.52 \\
    CFV2~\cite{cfv2}   & 86.26  & 86.53  & 16.90  & 10.10  & 88.40  & 60.53 \\
    TPN~\cite{tpn}   & 37.86  & 8.79  & 90.04  & 68.25  & 97.34  & 61.52 \\
    TIN~\cite{tin}   & 33.78  & 21.47  & 79.44  & 81.59  & 97.88  & 63.97 \\
    TRN~\cite{trn}   & 38.92  & 26.64  & 93.98  & 91.34  & 93.97  & 71.26 \\
    TSM~\cite{tsm}   & 70.37  & 54.70  & 70.37  & 78.46  & 96.76  & 76.40 \\
    X3D~\cite{x3d}   & 92.39  & 37.27  & 49.60  & 65.72  & 97.51  & 77.09 \\
    TimeSformer~\cite{timesformer}   & 73.14  & 20.17  & 39.40  & 74.80  & 92.32  & 64.28 \\
    UniFormer V2~\cite{uniformerv2}   & 20.05  & 14.81  & 99.21  & 45.21  & 96.89  & 57.55 \\
    VideoSwin~\cite{videoswin}   & 62.29  & 8.01  & 45.83  & 91.82  & 99.29  & 67.27 \\
    MViT V2~\cite{mvit}   & 76.34  & 98.29  & 96.62  & 47.50  & 97.58  & 79.90 \\
    ESSP~\cite{chen2024single}  &2.44 &6.80 &5.33 &0.15 &98.16 &22.83 \\
    LOTA~\cite{lota2025}  &77.65 &63.31 &79.57 &79.68 &76.12 &75.32 \\
   \midrule
    NAMP + I3D  & \textbf{92.20}  & 80.60  & \textbf{98.71}  & \textbf{96.76}  & 90.81  & \textbf{91.92} \\
    NAMP + SlowFast & 86.92  & 94.56  & 88.14  & 93.36  & 87.81  & 90.18 \\
    NAMP + Timesformer  & 98.05  & 96.55  & 100.0  & 100.0  & 3.14  & 79.32 \\
   \bottomrule
    \end{tabular}}
    \caption{\textbf{Comparison with state-of-art methods on the GenVidBench.} The GenVidBench has five subsets, and Avg. denotes the averaged accuracy. } 
  \label{comparison}
\end{table}

\begin{table}[tbp]
  \centering
  \resizebox{0.5\textwidth}{!}{
    \begin{tabular}{l|c|c|c|c|c|c}
    \toprule
    \textbf{Method} & \textbf{COG} & \textbf{T2VZ} & \textbf{TAV} & \textbf{VC} & \textbf{YT-BI} & \textbf{Avg.} \\
   \midrule
     I3D~\cite{i3d} & 57.40  & 59.40  & 31.40  & 48.40  & 52.44  & 50.80 \\
     SlowFast~\cite{slowfast} & 21.00  & 1.40  & 52.80  & 6.80  & 39.60  & 30.05 \\
     ESSP~\cite{chen2024single} & 70.00  & 72.40  & 60.80  & 60.40  & 13.70  & 39.80 \\
    \midrule
     NAMP w/o Patch  & 39.60  & 25.20  & 50.00  & 41.00  & 98.55  & 68.74 \\
     NAMP w/o Noise & 23.80  & 41.20  & 49.20  & 50.80  & 91.10  & 66.18 \\  
     NAMP (ours)  & 42.60  & 84.60  & 90.60  & 89.00  & 63.20  & 69.95 \\
   \bottomrule
    \end{tabular}}
  \caption{\textbf{Cross-dataset AI-generated video detection on the HardGVD. } The averaged accuracy is more reliable than the accuracies of the subsets.} 
  \label{hardGVD}
\end{table}

\begin{table}[tbp]
  
  \centering
  \resizebox{0.3\textwidth}{!}{
    \begin{tabular}{l|c|c}
    \toprule
    \textbf{Method} & \textbf{WANX} & \textbf{LTX-Video}  \\
\midrule
I3D~\cite{i3d}	 & 30.6   & 42.00 \\
SlowFast~\cite{slowfast}   & 27.67	 & 44.33 \\
Timesformer~\cite{timesformer}	 & 50.33	 & 55.33 \\
\midrule
NAMP (ours)	& 84.00	& 75.33 \\
\bottomrule
    \end{tabular}}
    \caption{\textbf{Evaluation on videos generated by new video generation models.} We generate 300 videos for each model and directly evaluate the models trained on GenVidBench using these subsets.} 
  \label{tab:cross_eval}
\end{table}

\noindent\textbf{Comparison on the GenVidBench: } We apply our methods on several video classification methods, including I3D~\cite{i3d}, SlowFast~\cite{slowfast} and TimeSformer~\cite{timesformer}, and compare them with state-of-the-art (SOTA) methods, including several mainstream deepfake image detection methods (ESSP~\cite{chen2024single}, LOTA~\cite{lota2025}). The results are shown in Table~\ref{comparison}.
Across nearly all benchmarks, our method significantly outperforms SOTA methods. For example, in terms of averaged accuracy, our approach using I3D surpasses SOTA methods by a margin of 12.02\%. Moreover, other SOTA methods often struggle to achieve balanced performance across all subsets, typically performing well on some while showing a noticeable decline on others. In contrast, our method consistently achieves a competitive performance exceeding 80\% across all subsets. Besides, due to the fundamentally different mechanisms, a substantial domain gap exists between AI-generated images and AI-generated videos, which is much larger than the gap among videos by different generators~\cite{vahdati2024beyond}. Thus, methods designed for AI image detection also do not generalize well on AI videos (ESSP~\cite{chen2024single}, LOTA~\cite{lota2025}). 

\noindent\textbf{Cross-dataset evaluation on the HardGVD:} We directly evaluate the model trained on the GenVidBench dataset on the challenging HardGVD dataset. 
In Table~\ref{hardGVD}, we establish important baselines on this dataset, including I3D~\cite{i3d}, SlowFast~\cite{slowfast}, and ESSP~\cite{chen2024single}. Using I3D~\cite{i3d} as the discriminator, we evaluate different variants of the proposed NAMP. NAMP w/o Patch refers to the variant that uses the noise frame without spatial amplification, while NAMP w/o Noise denotes the variant that directly performs spatial amplification and temporal aggregation on the RGB video. We find that existing SOTA methods are largely ineffective, but our method still achieves nearly 70\% accuracy.

\noindent\textbf{Generalization to Advanced Video Generators: } We use new video generation models, namely WANX~\cite{wang2025wan} and LTX Video~\cite{hacohen2024ltx}, to create new evaluation subsets. Both subsets contain 500 videos that are long (120 frames per video) and high-resolution (1280$\times$720 pixels). We then evaluate our model on these subsets, and the results are presented in Table~\ref{tab:cross_eval}. 
The results demonstrate that while alternative methods generally achieve poor performance (below 60\% accuracy) on these new subsets, our proposed approach maintains superior effectiveness. These experiments underscores the generalization capability and robustness of our method.

\label{sec:ablation}
\subsection{Ablation Studies and Analyses}

\noindent \textbf{Ablation Studies: }The key components of our approach are bit-plane-based noise extraction and divergence-based patch selection. We perform ablation studies on these components in Table~\ref{ablation}. We find that removing patch selection and amplification causes the averaged accuracy to drop significantly from 91.92\% to 53.71\%. Furthermore, omitting noise extraction results in an accuracy decrease of 25.18\%. When both components are removed, the baseline exhibits a 15.29\% performance drop compared to the variant using noise alone. These results provide strong evidence for the effectiveness of the components of our approach.

\begin{table*}[ht]
    \centering
    \begin{minipage}[t]{\columnwidth} 
        \centering
        \adjustbox{max width=\textwidth}{
        \begin{tabular}{cc|c|c|c|c|c|c}
    \toprule
    \textbf{Noise} & \textbf{Patch} & \textbf{MuseV} & \textbf{SVD} & \textbf{Mora} & \textbf{CogVideo} & \textbf{HD-VG} & \textbf{Avg.} \\
   \midrule
    &           & 15.54  & 13.66  & 30.99  & 34.60  & 93.99  & 38.10 \\
    \ding{51} &           & 17.88   & 62.15   & 45.17   & 48.12   & 92.61  & 53.39 \\
              & \ding{51} & 66.00  & 48.25  & 77.69  & 56.72  & 4.25  & 66.74 \\
    \ding{51} & \ding{51} & 92.20  & 80.60  & 98.71  & 96.76  & 90.81  & 91.92  \\
     \bottomrule
    \end{tabular}}
    \vspace{-0.2cm}
    \caption{\textbf{Ablation studies on the GenVidBench.} Noise denotes the bit-plane-based noise extraction, and patch refers to the process of patch selection and spatial amplification.} 
  \label{ablation}
    \end{minipage}
    \hfill
    \begin{minipage}[t]{\columnwidth} 
        \centering
        \adjustbox{max width=\textwidth}{
        \begin{tabular}{cc|c|c|c|c|c|c}
    \toprule
    \textbf{Patch Section} & \textbf{Aggregation} & \textbf{MuseV} & \textbf{SVD} & \textbf{Mora} & \textbf{CogVideo} & \textbf{HD-VG} & \textbf{Avg.} \\
   \midrule
    Maximum  & Voxel-based  & 92.20  & 80.60  & 98.71  & 96.76  & 90.81  & 91.92 \\
    Minimum  & Voxel-based  &  93.17  & 97.31  & 96.61  & 7.14  & 57.14  & 69.94 \\
    Random   & Voxel-based  & 92.67  & 96.26  & 90.12  & 59.24  & 37.44  & 74.89 \\
   \midrule
    Maximum  & Frame-wise  & 90.28  & 96.93  & 89.47  & 93.36  & 61.58  & 86.25 \\
    Minimum &  Frame-wise  & 71.88  & 90.50  & 83.41  & 7.94  & 87.68  & 68.12 \\
    Random  &  Frame-wise  & 83.04  & 88.50  & 84.08  & 56.31  & 54.19  & 73.07 \\
    \bottomrule
    \end{tabular}}
    \vspace{-0.2cm}
    \caption{Impacts of patch selection and aggregation for AI-generated video detection.} 
  \label{selection}
    \end{minipage}
    \hfill
    \begin{minipage}[t]{\columnwidth} 
        \vspace{0.2cm}
        \centering
        \adjustbox{max width=\textwidth}{
        \begin{tabular}{l|ccccc|c}
    \toprule
    \small \textbf{Texture} & \textbf{MuseV} & \textbf{SVD} & \textbf{Mora} & \textbf{CogVideo} & \textbf{HD-VG} & \textbf{Avg.} \\
    \midrule
    \small Ours & 92.20 & 80.60 & 98.71 & 96.76 & 90.81 & 91.92\\
    \small High-divergence & 90.91 & 81.17 & 94.57 & 92.12 & 90.00 & 89.79\\
    \small Low-divergence & 91.92 & 79.77 & 95.86 & 94.54 & 92.34 & 90.93\\
    \toprule
    \end{tabular}
   }
   \vspace{-0.2cm}
   \caption{Impacts of different spatial texture-based patch selection methods, including high-divergence based, low-divergence based and ours.} 
   \label{texture-based}
    \end{minipage}
    \hfill
    \begin{minipage}[t]{\columnwidth} 
        \vspace{0.2cm}
        \centering
        \adjustbox{max width=\textwidth}{
    \begin{tabular}{l|ccccc|c}
    \toprule
    \small Method & MuseV & SVD & Mora & CogVideo & HD-VG & Avg. \\
    \midrule
    \small Motion-based & 81.08 & 98.28 & 81.24 & 82.68 & 65.67 & 81.66\\
    \small Flow-based & 81.45 & 98.20 & 81.89 & 83.91 & 70.59 & 83.10\\ 
    \small Gradient-based & \textbf{92.20} & \textbf{80.60} & \textbf{98.71} & \textbf{96.76} & \textbf{90.81} & \textbf{91.92}\\
    \toprule
    \end{tabular}
   }
   \vspace{-0.2cm}
   \caption{Impacts of different temporal motion-based patch selection methods, including motion based, flow based and gradient based.} 
  \label{motion-based}
    \end{minipage}
    \hfill
    \begin{minipage}[t]{\columnwidth} 
        \vspace{0.2cm}
        \centering
        \adjustbox{max width=\textwidth}{
        \begin{tabular}{c|c|c|c|c|c|c}
    \toprule
    \textbf{Bit-Planes} & \textbf{MuseV} & \textbf{SVD} & \textbf{Mora} & \textbf{CogVideo} & \textbf{HD-VG} & \textbf{Avg.} \\
   \midrule
    0 Bits  & 29.67  & 48.87  & 41.64  & 67.61  & 83.06  & 54.40 \\
    0$\sim$1 Bits  & 12.25  & 48.26  & 30.40  & 35.99  & 94.12  & 44.42 \\
    0$\sim$2 Bits  & 92.20  & 80.60  & 98.71  & 96.76  & 90.81  & 91.92 \\
    0$\sim$3 Bits  & 95.33  & 98.99  & 95.22  & 96.28  & 44.35  & 85.90 \\
    \bottomrule
    \end{tabular}}
    \vspace{-0.2cm}
    \caption{\textbf{Impacts of number of bit-planes.} Each RGB frame contains eight bit-planes, numbered from 0 to 7, where lower bit-planes represent finer details or noise. We use varying numbers of lower bit-planes to generate the noise sequence, and compare the results of AI-generated video detection.} 
  \label{plane}
    \end{minipage}
    \hfill
    \begin{minipage}[t]{\columnwidth} 
        \vspace{0.2cm}
        \centering
        \adjustbox{max width=\textwidth}{
        \begin{tabular}{c|c|c|c|c|c|c}
    \toprule
\textbf{Patch size} & \textbf{MuseV} & \textbf{SVD} & \textbf{Mora} & \textbf{CogVideo} & \textbf{HD-VG} & \textbf{Avg.} \\
   \midrule
    8$\times$8   & 2.86 & 2.03 & 1.93 & 0 & 99.01 & 21.44 \\
    16$\times$16 & 92.20  & 80.60  & 98.71  & 96.76  & 90.81  & 91.92 \\
    24$\times$24 & 99.60 & 99.78 & 97.22 & 99.88 & 16.80 & 82.44 \\
    32$\times$32 & 99.65 & 99.66 & 99.83 & 99.61 & 12.69 & 82.07 \\
   \bottomrule
    \end{tabular}}
    \vspace{-0.2cm}
    \caption{\textbf{Impacts of patch size in the spatial amplification.} We choose different patch sizes, i.e., 8$\times$8, 16$\times$16, 32$\times$32, and 64$\times$64, in the spatial amplification module, and compare the corresponding fake detection results. The default choice of 16$\times$16 performs more generalizable than the others.} 
  \label{size}
    \end{minipage}
\end{table*}

\noindent \textbf{Impacts of Patch Selection and Aggregation:}
In previous section, we select the patch with the maximum score and present voxel-based and frame-wise method for temporal aggregation. To verify the effectiveness of patch selection, we apply three patch selection strategies during testing: maximum, minimum, and random. We also compare two temporal aggregation methods: voxel-based and frame-wise. Table~\ref{selection} investigates the impacts of patch selection and aggregation. The results show that the maximum selection strategy achieves significantly better performance than the minimum and random selection strategies, exceeding them by 21.98\% and 17.03\%, respectively. In addition, the voxel-based aggregation, which selects the spatiotemporal region of the video and well preserves temporal dynamics, also consistently outperforms the frame-wise aggregation. The results demonstrate the effectiveness of our patch selection and aggregation strategy, and highlight the importance of temporal aggregation modules and temporal dynamics information for AI-generated video detection. Besides, we also select patches with either rich or uniform textures, as well as different temporal-based patch selection strategies such as motion-based and flow-based methods. Although our strategy performs considerably better, the other strategies using our bit-plane based methods also outperform the SOTA baselines. As a result, our approach is robust to different patch selection strategies.

\begin{table}[tbp]
  \centering
  \resizebox{0.5\textwidth}{!}{
     \begin{tabular}{l|c|c|c|c|c|c}
    \toprule
    \textbf{Method} & \textbf{MuseV} & \textbf{SVD} & \textbf{Mora} & \textbf{CogVideo} & \textbf{HD-VG} & \textbf{Avg.} \\
   \midrule
NAMP (ours) & 92.20 & 80.60 & 98.71 & 96.73 & 90.81 & 91.92 \\
\midrule
+ JPEG95 & 87.56 & 73.41 & 92.46 & 90.72 & 88.71 & 86.66 \\
+ JPEG90 & 82.39 & 70.05 & 85.19 & 85.62 & 86.86 & 82.10 \\
+ JPEG85 & 80.54 & 68.57 & 82.80 & 83.97 & 84.49 & 80.15 \\
\midrule
+ Gauss1 & 88.35 & 74.46 & 96.25 & 93.53 & 88.77 & 88.37 \\
+ Gauss2 & 85.34 & 74.75 & 95.26 & 92.33 & 86.37 & 86.90 \\
+ Gauss3 & 84.87 & 73.64 & 95.12 & 91.44 & 85.26 & 84.21 \\
\midrule
+ CRF-18 & 90.46 & 77.26 & 95.93 & 94.37 & 82.55 & 88.17 \\
+ CRF-24 & 83.36 & 71.75 & 88.37 & 87.55 & 71.42 & 80.53 \\
+ CRF-30 & 74.95 & 63.87 & 85.48 & 83.54 & 62.62 & 74.16 \\
\bottomrule
    \end{tabular}}
  \caption{\textbf{Robustness evaluation on perturbed videos.} Our model is evaluated on the GenVidBench dataset with JPEG compression (quality levels of 95\%, 90\%, and 85\%), Gaussian blur (standard deviation $\sigma$ = 1, 2, and 3), and H.264 comprssion (CRF-18, 24, 30). } 
  \label{tab:pertubation}
\end{table}

\begin{figure*}[t]
    \centering
    \includegraphics[width=0.95\linewidth]{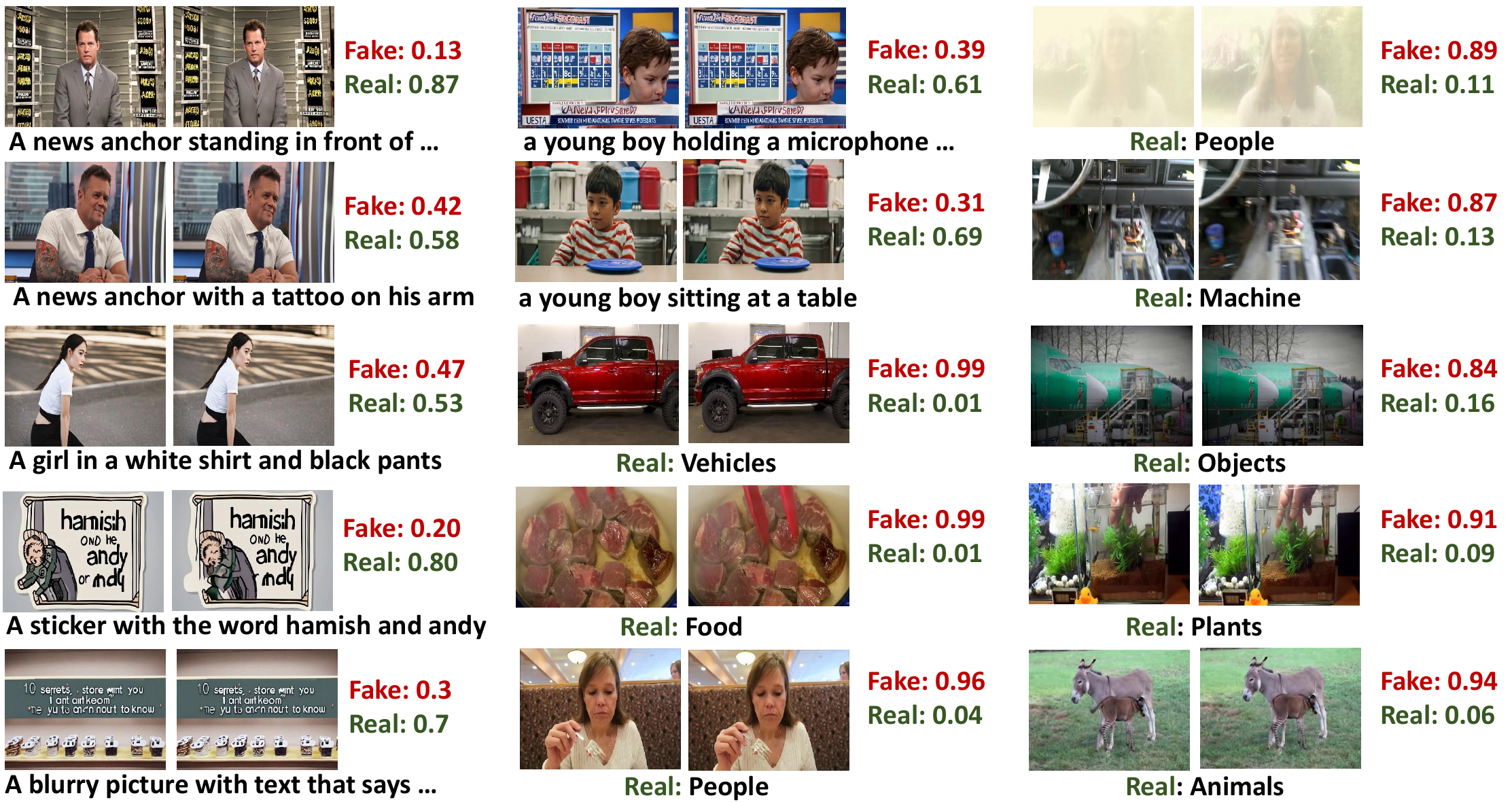}
    \caption{\textbf{Visualization results of hard examples for AI-generated video detection.} We visualize the hard examples on the GenVidBench that are misclassified by our approach for both AI-generated and real videos. Real videos are labeled as \textit{real} in the text below each video, while textual prompts for AI-generated videos are also displayed beneath the corresponding videos.}
    \label{fig:vis}
\end{figure*}

\noindent \textbf{Impacts of Number of Bit-Planes:} 
To investigate impacts of the number of bit-planes for AI-generated video detection, we choose the least bit-plane, 0$\sim$1 bit-planes, 0$\sim$2 bit-planes and 0$\sim$3 bit-planes for training and testing, respectively. The results are summarized in Table~\ref{plane}. From observations, generated artifacts mostly appear in low-bit planes (0$\sim$2 bit-planes), although 0$\sim$3 bit-planes significantly outperforms 0-1 bit-planes due to more information, it is 6$\%$ lower 
than 0$\sim$2 bit-planes as bit-3 contains more video detail than artifacts. We observe that using 0$\sim$2 bit-planes performs the best, and achieves satisfactory results on all subsets. This suggests that using too few number of bit-planes provides insufficient information for fake detection, whereas using too many may introduce irrelevant visual content information that hinders fake detection. 

\noindent \textbf{Impacts of Patch Size: }
Patch size is also an important hyperparameter in the spatial amplification module in Sec.~\ref{sec:patch_select}, as large patch sizes result in less spatial amplification of artifacts, while small patch sizes contain less artifact information. Table~\ref{size} shows the impact of patch size on AI-generated video detection in our approach. We find that when using small patch sizes, the model tends to classify all videos as real, with an average accuracy of 21.44$\%$ (patch size of 8$\times$8). In contrast, with large patch sizes, it tends to classify all videos as fake, with an average accuracy of 82.07$\%$ (patch size of 32$\times$32). The best patch size is 16$\times$16, which achieves excellent performance on most subsets and a superb average accuracy of 91.92$\%$.

\noindent \textbf{Robustness to Video Degradation: }
To further demonstrate robustness to perturbations and degradations, we test the model on video frames affected by varying levels of JPEG compression and Gaussian blur, and results are given in Table~\ref{tab:pertubation}. Our model maintains strong detection performance on these degraded videos, with accuracy dropping by no more than 12\% and 8\% in both cases, demonstrating the robustness of our approach. Moreover, we subject the videos in GenVidBench to H.264 compression, incrementally increasing the CRF value, and our NAMP maintains an accuracy of 74.16\% (dropping by no more than 16\% ) even at a CRF of 30, demonstrating strong robustness to video degradation.

\noindent \textbf{Visualizations:} To better understand the difficulty of the task of AI-generated video detection and the weaknesses of our model, we visualize hard examples that are misclassified by our approach for both AI-generated and real videos in Figure~\ref{fig:vis}. We find that most of the AI-generated videos misclassified as real by our model are static person, objects and backgrounds that contain text, while real videos misclassified as fake by our model tend to have blurry content or plain yet textured backgrounds. The hard samples of videos mainly consist of static visual content and lack motion features and temporal dynamics. These results further indicate the advantages of our model in detecting AI-generated videos with temporal dynamics.

\section{Conclusion and Limitations}
\label{sec:conclusion}
We address AI-generated video detection from the novel perspective of bit-planes, and propose a straightforward yet effective approach. To facilitate the training of the discriminator networks using the noise sequences extracted from bit-planes, our approach employs three modules, i.e., intensity enhancement, spatial amplification and temporal aggregation, to amplify these noises. Our method significantly surpasses existing approaches on the GenVidBench benchmark, boosting the average video fake detection accuracy to more than 90\%. To facilitate progress in this field, we curate a challenging benchmark dataset named the HardGVD through comprehensive data collection, and conduct evaluations of several representative methods on this benchmark. One limitation of our approach lies in the instability of training outcomes, which might be attributed to the random sampling of image patches and frames from the videos. In the future, we will design a more effective artifact region selection strategy to mitigate training instability.

\section*{Acknowledgments}
This work was supported by National Science Foundation of China (62302093, 52441503), Jiangsu Province Natural Science Fund (BK20230833), the CIPS-SMP-Zhipu Large Model Fund, the Fundamental Research Funds for the Central Universities (2242025K30024), the Open Research Fund of the State Key Laboratory of Multimodal Artificial Intelligence Systems (E5SP060116) and the Big Data Computing Center of Southeast University.

{
    \small
    \bibliographystyle{ieeenat}
    \bibliography{main}

\begin{thebibliography}{86}
\providecommand{\natexlab}[1]{#1}
\providecommand{\url}[1]{\texttt{#1}}
\expandafter\ifx\csname urlstyle\endcsname\relax
  \providecommand{\doi}[1]{doi: #1}\else
  \providecommand{\doi}{doi: \begingroup \urlstyle{rm}\Url}\fi

\bibitem[Bai et~al.(2024)Bai, Lin, Cao, and Lou]{bai2024ai}
Jianfa Bai, Man Lin, Gang Cao, and Zijie Lou.
\newblock Ai-generated video detection via spatial-temporal anomaly learning.
\newblock In \emph{Chinese Conference on Pattern Recognition and Computer Vision}, pages 460--470. Springer, 2024.

\bibitem[Bertasius et~al.(2021)Bertasius, Wang, and Torresani]{timesformer}
Gedas Bertasius, Heng Wang, and Lorenzo Torresani.
\newblock Is space-time attention all you need for video understanding?, 2021.

\bibitem[Berthouzoz et~al.(2011)Berthouzoz, Li, Dontcheva, and Agrawala]{berthouzoz2011framework}
Floraine Berthouzoz, Wilmot Li, Mira Dontcheva, and Maneesh Agrawala.
\newblock A framework for content-adaptive photo manipulation macros: Application to face, landscape, and global manipulations.
\newblock \emph{ACM Transactions on Graphics}, 30\penalty0 (5):\penalty0 1--14, 2011.

\bibitem[Blattmann et~al.(2023{\natexlab{a}})Blattmann, Dockhorn, Kulal, Mendelevitch, Kilian, Lorenz, Levi, English, Voleti, Letts, Jampani, and Rombach]{svd}
Andreas Blattmann, Tim Dockhorn, Sumith Kulal, Daniel Mendelevitch, Maciej Kilian, Dominik Lorenz, Yam Levi, Zion English, Vikram Voleti, Adam Letts, Varun Jampani, and Robin Rombach.
\newblock Stable video diffusion: Scaling latent video diffusion models to large datasets, 2023{\natexlab{a}}.

\bibitem[Blattmann et~al.(2023{\natexlab{b}})Blattmann, Dockhorn, Kulal, Mendelevitch, Kilian, Lorenz, Levi, English, Voleti, Letts, et~al.]{blattmann2023stable}
Andreas Blattmann, Tim Dockhorn, Sumith Kulal, Daniel Mendelevitch, Maciej Kilian, Dominik Lorenz, Yam Levi, Zion English, Vikram Voleti, Adam Letts, et~al.
\newblock Stable video diffusion: Scaling latent video diffusion models to large datasets.
\newblock \emph{arXiv preprint arXiv:2311.15127}, 2023{\natexlab{b}}.

\bibitem[Bohacek and Farid(2024)]{bohacek2024lost}
Matyas Bohacek and Hany Farid.
\newblock Lost in translation: Lip-sync deepfake detection from audio-video mismatch.
\newblock In \emph{Proceedings of the IEEE/CVF Conference on Computer Vision and Pattern Recognition}, pages 4315--4323, 2024.

\bibitem[Carreira and Zisserman(2017)]{i3d}
J. Carreira and Andrew Zisserman.
\newblock Quo vadis, action recognition? a new model and the kinetics dataset.
\newblock In \emph{IEEE/CVF Conference on Computer Vision and Pattern Recognition}, pages 4724--4733, 2017.

\bibitem[Chen et~al.(2024{\natexlab{a}})Chen, Hong, Huang, Xu, Gu, Li, Lan, Zhu, Zhang, Wang, et~al.]{demamba}
Haoxing Chen, Yan Hong, Zizheng Huang, Zhuoer Xu, Zhangxuan Gu, Yaohui Li, Jun Lan, Huijia Zhu, Jianfu Zhang, Weiqiang Wang, et~al.
\newblock Demamba: Ai-generated video detection on million-scale genvideo benchmark.
\newblock \emph{arXiv preprint arXiv:2405.19707}, 2024{\natexlab{a}}.

\bibitem[Chen et~al.(2024{\natexlab{b}})Chen, Zhang, Cun, Xia, Wang, Weng, and Shan]{vc2}
Haoxin Chen, Yong Zhang, Xiaodong Cun, Menghan Xia, Xintao Wang, Chao Weng, and Ying Shan.
\newblock Videocrafter2: Overcoming data limitations for high-quality video diffusion models.
\newblock \emph{arXiv preprint arXiv:2401.09047}, 2024{\natexlab{b}}.

\bibitem[Chen et~al.(2024{\natexlab{c}})Chen, Yao, and Niu]{chen2024single}
Jiaxuan Chen, Jieteng Yao, and Li Niu.
\newblock A single simple patch is all you need for ai-generated image detection.
\newblock \emph{arXiv preprint arXiv:2402.01123}, 2024{\natexlab{c}}.

\bibitem[Chen et~al.(2023)Chen, Wang, Zhang, Zhuang, Ma, Yu, Wang, Lin, Qiao, and Liu]{chen2023seine}
Xinyuan Chen, Yaohui Wang, Lingjun Zhang, Shaobin Zhuang, Xin Ma, Jiashuo Yu, Yali Wang, Dahua Lin, Yu Qiao, and Ziwei Liu.
\newblock Seine: Short-to-long video diffusion model for generative transition and prediction.
\newblock In \emph{The Twelfth International Conference on Learning Representations}, 2023.

\bibitem[Choi et~al.(2024)Choi, Kim, Jeong, Baek, and Choi]{choi2024exploiting}
Jongwook Choi, Taehoon Kim, Yonghyun Jeong, Seungryul Baek, and Jongwon Choi.
\newblock Exploiting style latent flows for generalizing deepfake video detection.
\newblock In \emph{Proceedings of the IEEE/CVF Conference on Computer Vision and Pattern Recognition}, pages 1133--1143, 2024.

\bibitem[Corvi et~al.(2025)Corvi, Cozzolino, Prashnani, De~Mello, Nagano, and Verdoliva]{corvi2025seeing}
Riccardo Corvi, Davide Cozzolino, Ekta Prashnani, Shalini De~Mello, Koki Nagano, and Luisa Verdoliva.
\newblock Seeing what matters: Generalizable ai-generated video detection with forensic-oriented augmentation.
\newblock In \emph{Annual Conference on Neural Information Processing Systems}, 2025.

\bibitem[Cozzolino et~al.(2024)Cozzolino, Poggi, Nie{\ss}ner, and Verdoliva]{cozzolino2024zero}
Davide Cozzolino, Giovanni Poggi, Matthias Nie{\ss}ner, and Luisa Verdoliva.
\newblock Zero-shot detection of ai-generated images.
\newblock In \emph{European Conference on Computer Vision}, pages 54--72. Springer, 2024.

\bibitem[Dolhansky et~al.(2020)Dolhansky, Bitton, Pflaum, Lu, Howes, Wang, and Ferrer]{dolhansky2020deepfake}
Brian Dolhansky, Joanna Bitton, Ben Pflaum, Jikuo Lu, Russ Howes, Menglin Wang, and Cristian~Canton Ferrer.
\newblock The deepfake detection challenge (dfdc) dataset.
\newblock \emph{arXiv preprint arXiv:2006.07397}, 2020.

\bibitem[Feichtenhofer(2020)]{x3d}
Christoph Feichtenhofer.
\newblock X3d: Expanding architectures for efficient video recognition, 2020.

\bibitem[Feichtenhofer et~al.(2019)Feichtenhofer, Fan, Malik, and He]{slowfast}
Christoph Feichtenhofer, Haoqi Fan, Jitendra Malik, and Kaiming He.
\newblock Slowfast networks for video recognition.
\newblock In \emph{Proceedings of the IEEE International Conference on Computer Vision}, pages 6202--6211, 2019.

\bibitem[Frank et~al.(2020)Frank, Eisenhofer, Sch{\"o}nherr, Fischer, Kolossa, and Holz]{frank2020leveraging}
Joel Frank, Thorsten Eisenhofer, Lea Sch{\"o}nherr, Asja Fischer, Dorothea Kolossa, and Thorsten Holz.
\newblock Leveraging frequency analysis for deep fake image recognition.
\newblock In \emph{International Conference on Machine Learning}, pages 3247--3258. PMLR, 2020.

\bibitem[Goodfellow et~al.(2014)Goodfellow, Pouget-Abadie, Mirza, Xu, Warde-Farley, Ozair, Courville, and Bengio]{goodfellow2014generative}
Ian Goodfellow, Jean Pouget-Abadie, Mehdi Mirza, Bing Xu, David Warde-Farley, Sherjil Ozair, Aaron Courville, and Yoshua Bengio.
\newblock Generative adversarial nets.
\newblock In \emph{Conference on Neural Information Processing Systems}, 2014.

\bibitem[G{\"u}era and Delp(2018)]{guera2018deepfake}
David G{\"u}era and Edward~J Delp.
\newblock Deepfake video detection using recurrent neural networks.
\newblock In \emph{IEEE International Conference on Advanced Video and Signal Based Surveillance (AVSS)}, pages 1--6, 2018.

\bibitem[Guo et~al.(2024)Guo, Yang, Rao, Liang, Wang, Qiao, Agrawala, Lin, and Dai]{Animatediff}
Yuwei Guo, Ceyuan Yang, Anyi Rao, Zhengyang Liang, Yaohui Wang, Yu Qiao, Maneesh Agrawala, Dahua Lin, and Bo Dai.
\newblock {AnimateDiff: Animate Your Personalized Text-to-Image Diffusion Models without Specific Tuning}.
\newblock In \emph{International Conference on Learning Representations}, 2024.

\bibitem[HaCohen et~al.(2024{\natexlab{a}})HaCohen, Chiprut, Brazowski, Shalem, Moshe, Richardson, Levin, Shiran, Zabari, Gordon, et~al.]{LTX}
Yoav HaCohen, Nisan Chiprut, Benny Brazowski, Daniel Shalem, Dudu Moshe, Eitan Richardson, Eran Levin, Guy Shiran, Nir Zabari, Ori Gordon, et~al.
\newblock {LTX-Video: Realtime Video Latent Diffusion}.
\newblock \emph{arXiv preprint arXiv:2501.00103}, 2024{\natexlab{a}}.

\bibitem[HaCohen et~al.(2024{\natexlab{b}})HaCohen, Chiprut, Brazowski, Shalem, Moshe, Richardson, Levin, Shiran, Zabari, Gordon, et~al.]{hacohen2024ltx}
Yoav HaCohen, Nisan Chiprut, Benny Brazowski, Daniel Shalem, Dudu Moshe, Eitan Richardson, Eran Levin, Guy Shiran, Nir Zabari, Ori Gordon, et~al.
\newblock {LTX-Video: R}ealtime video latent diffusion.
\newblock \emph{arXiv preprint arXiv:2501.00103}, 2024{\natexlab{b}}.

\bibitem[Hong et~al.(2022)Hong, Ding, Zheng, Liu, and Tang]{cogvideo}
Wenyi Hong, Ming Ding, Wendi Zheng, Xinghan Liu, and Jie Tang.
\newblock Cogvideo: Large-scale pretraining for text-to-video generation via transformers.
\newblock \emph{arXiv preprint arXiv:2205.15868}, 2022.

\bibitem[Hsu et~al.(2024)Hsu, Chen, Wu, Wang, Lee, and Chou]{hsu2024grace}
Chih-Chung Hsu, Shao-Ning Chen, Mei-Hsuan Wu, Yi-Fang Wang, Chia-Ming Lee, and Yi-Shiuan Chou.
\newblock Grace: Graph-regularized attentive convolutional entanglement with laplacian smoothing for robust deepfake video detection.
\newblock \emph{arXiv preprint arXiv:2406.19941}, 2024.

\bibitem[Intern{\`o} et~al.(2025)Intern{\`o}, Geirhos, Olhofer, Liu, Hammer, and Klindt]{interno2025ai}
Christian Intern{\`o}, Robert Geirhos, Markus Olhofer, Sunny Liu, Barbara Hammer, and David Klindt.
\newblock Ai-generated video detection via perceptual straightening.
\newblock In \emph{Annual Conference on Neural Information Processing Systems}, 2025.

\bibitem[Juefei-Xu et~al.(2022)Juefei-Xu, Wang, Huang, Guo, Ma, and Liu]{juefei2022countering}
Felix Juefei-Xu, Run Wang, Yihao Huang, Qing Guo, Lei Ma, and Yang Liu.
\newblock Countering malicious deepfakes: Survey, battleground, and horizon.
\newblock \emph{International Journal of Computer Vision}, 130\penalty0 (7):\penalty0 1678--1734, 2022.

\bibitem[Karamchand(2025)]{misuse-1}
Gopalakrishna Karamchand.
\newblock {Detecting the Abuse of Generative AI in Cybersecurity Contexts: Challenges, Frameworks, and Solutions}.
\newblock \emph{Journal of Data Analysis and Critical Management}, 1\penalty0 (03):\penalty0 1--12, 2025.

\bibitem[Kaur et~al.(2024)Kaur, Noori~Hoshyar, Saikrishna, Firmin, and Xia]{kaur2024deepfake}
Achhardeep Kaur, Azadeh Noori~Hoshyar, Vidya Saikrishna, Selena Firmin, and Feng Xia.
\newblock Deepfake video detection: challenges and opportunities.
\newblock \emph{Artificial Intelligence Review}, 57\penalty0 (6):\penalty0 159, 2024.

\bibitem[Khachatryan et~al.(2023{\natexlab{a}})Khachatryan, Movsisyan, Tadevosyan, Henschel, Wang, Navasardyan, and Shi]{t2vz}
Levon Khachatryan, Andranik Movsisyan, Vahram Tadevosyan, Roberto Henschel, Zhangyang Wang, Shant Navasardyan, and Humphrey Shi.
\newblock Text2video-zero: Text-to-image diffusion models are zero-shot video generators.
\newblock In \emph{Proceedings of the IEEE/CVF International Conference on Computer Vision}, pages 15954--15964, 2023{\natexlab{a}}.

\bibitem[Khachatryan et~al.(2023{\natexlab{b}})Khachatryan, Movsisyan, Tadevosyan, Henschel, Wang, Navasardyan, and Shi]{text2video-zero}
Levon Khachatryan, Andranik Movsisyan, Vahram Tadevosyan, Roberto Henschel, Zhangyang Wang, Shant Navasardyan, and Humphrey Shi.
\newblock Text2video-zero: Text-to-image diffusion models are zero-shot video generators.
\newblock \emph{arXiv preprint arXiv:2303.13439}, 2023{\natexlab{b}}.

\bibitem[Kong et~al.(2024)Kong, Tian, Zhang, Min, Dai, Zhou, Xiong, Li, Wu, Zhang, et~al.]{HYV}
Weijie Kong, Qi Tian, Zijian Zhang, Rox Min, Zuozhuo Dai, Jin Zhou, Jiangfeng Xiong, Xin Li, Bo Wu, Jianwei Zhang, et~al.
\newblock {HunyuanVideo: A Systematic Framework For Large Video Generative Models}.
\newblock \emph{arXiv preprint arXiv:2412.03603}, 2024.

\bibitem[Korshunov and Marcel(2018)]{korshunov2018deepfakes}
Pavel Korshunov and S{\'e}bastien Marcel.
\newblock Deepfakes: a new threat to face recognition? assessment and detection.
\newblock \emph{arXiv preprint arXiv:1812.08685}, 2018.

\bibitem[Korshunova et~al.(2017)Korshunova, Shi, Dambre, and Theis]{korshunova2017fast}
Iryna Korshunova, Wenzhe Shi, Joni Dambre, and Lucas Theis.
\newblock Fast face-swap using convolutional neural networks.
\newblock In \emph{Proceedings of the IEEE International Conference on Computer Vision}, pages 3677--3685, 2017.

\bibitem[Li et~al.(2022{\natexlab{a}})Li, Wang, He, Li, Wang, Wang, and Qiao]{uniformerv2}
Kunchang Li, Yali Wang, Yinan He, Yizhuo Li, Yi Wang, Limin Wang, and Yu Qiao.
\newblock Uniformerv2: Spatiotemporal learning by arming image vits with video uniformer.
\newblock \emph{arXiv preprint arXiv:2211.09552}, 2022{\natexlab{a}}.

\bibitem[Li and Lyu(2018)]{li2018exposing}
Yuezun Li and Siwei Lyu.
\newblock Exposing deepfake videos by detecting face warping artifacts.
\newblock \emph{arXiv preprint arXiv:1811.00656}, 2018.

\bibitem[Li et~al.(2020)Li, Yang, Sun, Qi, and Lyu]{li2020celeb}
Yuezun Li, Xin Yang, Pu Sun, Honggang Qi, and Siwei Lyu.
\newblock Celeb-df: A large-scale challenging dataset for deepfake forensics.
\newblock In \emph{Proceedings of the IEEE/CVF Conference on Computer Vision and Pattern Recognition}, pages 3207--3216, 2020.

\bibitem[Li et~al.(2022{\natexlab{b}})Li, Wu, Fan, Mangalam, Xiong, Malik, and Feichtenhofer]{mvit}
Yanghao Li, Chao-Yuan Wu, Haoqi Fan, Karttikeya Mangalam, Bo Xiong, Jitendra Malik, and Christoph Feichtenhofer.
\newblock Mvitv2: Improved multiscale vision transformers for classification and detection.
\newblock In \emph{IEEE/CVF Conference on Computer Vision and Pattern Recognition}, 2022{\natexlab{b}}.

\bibitem[Lin et~al.(2019)Lin, Gan, and Han]{tsm}
Ji Lin, Chuang Gan, and Song Han.
\newblock Tsm: Temporal shift module for efficient video understanding, 2019.

\bibitem[Liu et~al.(2024)Liu, Zhang, Li, Yan, Gao, Chen, Yuan, Huang, Sun, Gao, et~al.]{Sora}
Yixin Liu, Kai Zhang, Yuan Li, Zhiling Yan, Chujie Gao, Ruoxi Chen, Zhengqing Yuan, Yue Huang, Hanchi Sun, Jianfeng Gao, et~al.
\newblock {Sora: A Review on Background, Technology, Limitations, and Opportunities of Large Vision Models}.
\newblock \emph{arXiv preprint arXiv:2402.17177}, 2024.

\bibitem[Liu et~al.(2022)Liu, Ning, Cao, Wei, Zhang, Lin, and Hu]{videoswin}
Ze Liu, Jia Ning, Yue Cao, Yixuan Wei, Zheng Zhang, Stephen Lin, and Han Hu.
\newblock Video swin transformer.
\newblock In \emph{Proceedings of the IEEE/CVF Conference on Computer Vision and Pattern Recognition}, pages 3202--3211, 2022.

\bibitem[Luo et~al.(2024)Luo, Du, Yan, and Ding]{luo2024lare}
Yunpeng Luo, Junlong Du, Ke Yan, and Shouhong Ding.
\newblock Lare\^{} 2: Latent reconstruction error based method for diffusion-generated image detection.
\newblock In \emph{IEEE/CVF Conference on Computer Vision and Pattern Recognition}, pages 17006--17015, 2024.

\bibitem[Ma et~al.(2025)Ma, Huang, Yan, Chen, Duan, Yin, Wan, Ming, Song, Chen, et~al.]{StepVideo}
Guoqing Ma, Haoyang Huang, Kun Yan, Liangyu Chen, Nan Duan, Shengming Yin, Changyi Wan, Ranchen Ming, Xiaoniu Song, Xing Chen, et~al.
\newblock {Step-Video-T2V Technical Report: The Practice, Challenges, and Future of Video Foundation Model}.
\newblock \emph{arXiv preprint arXiv:2502.10248}, 2025.

\bibitem[Nguyen et~al.(2019)Nguyen, Yamagishi, and Echizen]{cfv2}
Huy~H. Nguyen, Junichi Yamagishi, and Isao Echizen.
\newblock Use of a capsule network to detect fake images and videos, 2019.

\bibitem[Ni et~al.(2025)Ni, Yan, Huang, Yuan, Tang, Hu, Chen, and Wang]{ni2025genvidbenchchallengingbenchmarkdetecting}
Zhenliang Ni, Qiangyu Yan, Mouxiao Huang, Tianning Yuan, Yehui Tang, Hailin Hu, Xinghao Chen, and Yunhe Wang.
\newblock Genvidbench: A challenging benchmark for detecting ai-generated video, 2025.

\bibitem[Pang et~al.(2024)Pang, Xiong, Zhang, and Wang]{misuse-2}
Yan Pang, Aiping Xiong, Yang Zhang, and Tianhao Wang.
\newblock {Towards Understanding Unsafe Video Generation}.
\newblock \emph{arXiv preprint arXiv:2407.12581}, 2024.

\bibitem[pika(2024)]{pika}
pika.
\newblock Pika.
\newblock Online, 2024.
\newblock Available:\url{https://pika.art/home}.

\bibitem[Qian et~al.(2020)Qian, Yin, Sheng, Chen, and Shao]{f3net}
Yuyang Qian, Guojun Yin, Lu Sheng, Zixuan Chen, and Jing Shao.
\newblock Thinking in frequency: Face forgery detection by mining frequency-aware clues, 2020.

\bibitem[Rombach et~al.(2022)Rombach, Blattmann, Lorenz, Esser, and Ommer]{rombach2022high}
Robin Rombach, Andreas Blattmann, Dominik Lorenz, Patrick Esser, and Bj{\"o}rn Ommer.
\newblock High-resolution image synthesis with latent diffusion models.
\newblock In \emph{IEEE/CVF Conference on Computer Vision and Pattern Recognition}, 2022.

\bibitem[Rossler et~al.(2019)Rossler, Cozzolino, Verdoliva, Riess, Thies, and Nie{\ss}ner]{rossler2019faceforensics++}
Andreas Rossler, Davide Cozzolino, Luisa Verdoliva, Christian Riess, Justus Thies, and Matthias Nie{\ss}ner.
\newblock Faceforensics++: Learning to detect manipulated facial images.
\newblock In \emph{Proceedings of the IEEE/CVF International Conference on Computer Vision}, pages 1--11, 2019.

\bibitem[Sabir et~al.(2019)Sabir, Cheng, Jaiswal, AbdAlmageed, Masi, and Natarajan]{sabir2019recurrent}
Ekraam Sabir, Jiaxin Cheng, Ayush Jaiswal, Wael AbdAlmageed, Iacopo Masi, and Prem Natarajan.
\newblock Recurrent convolutional strategies for face manipulation detection in videos.
\newblock \emph{Interfaces (GUI)}, 3\penalty0 (1):\penalty0 80--87, 2019.

\bibitem[Shao et~al.(2020)Shao, Qian, and Liu]{tin}
Hao Shao, Shengju Qian, and Yu Liu.
\newblock Temporal interlacing network.
\newblock \emph{AAAI}, 2020.

\bibitem[Shao et~al.(2023)Shao, Wu, and Liu]{shao2023detecting}
Rui Shao, Tianxing Wu, and Ziwei Liu.
\newblock Detecting and grounding multi-modal media manipulation.
\newblock In \emph{Proceedings of the IEEE/CVF Conference on Computer Vision and Pattern Recognition}, pages 6904--6913, 2023.

\bibitem[Singer et~al.(2022)Singer, Polyak, Hayes, Yin, An, Zhang, Hu, Yang, Ashual, Gafni, et~al.]{singer2022make}
Uriel Singer, Adam Polyak, Thomas Hayes, Xi Yin, Jie An, Songyang Zhang, Qiyuan Hu, Harry Yang, Oron Ashual, Oran Gafni, et~al.
\newblock Make-a-video: Text-to-video generation without text-video data.
\newblock \emph{arXiv preprint arXiv:2209.14792}, 2022.

\bibitem[Spotlight(2025)]{earthquake}
Spotlight.
\newblock {AI-generated footage falsely used to show 'aftermath' of Philippines earthquake}.
\newblock \emph{https://spotlight.ebu.ch/}, 2025.

\bibitem[Studio(2024)]{morphstudio}
Morph Studio.
\newblock Morph studio.
\newblock \url{https://www.morphstudio.com/}, 2024.

\bibitem[Tolosana et~al.(2020)Tolosana, Vera-Rodriguez, Fierrez, Morales, and Ortega-Garcia]{tolosana2020deepfakes}
Ruben Tolosana, Ruben Vera-Rodriguez, Julian Fierrez, Aythami Morales, and Javier Ortega-Garcia.
\newblock Deepfakes and beyond: A survey of face manipulation and fake detection.
\newblock \emph{Information Fusion}, 64:\penalty0 131--148, 2020.

\bibitem[University(2024)]{vript}
Shanghai Jiao~Tong University.
\newblock Vript: A video is worth thousands of words.
\newblock Online, 2024.
\newblock Available:\url{https://github.com/mutonix/Vript}.

\bibitem[Vahdati et~al.(2024)Vahdati, Nguyen, Azizpour, and Stamm]{vahdati2024beyond}
Danial~Samadi Vahdati, Tai~D Nguyen, Aref Azizpour, and Matthew~C Stamm.
\newblock Beyond deepfake images: Detecting ai-generated videos.
\newblock In \emph{Proceedings of the IEEE/CVF Conference on Computer Vision and Pattern Recognition}, pages 4397--4408, 2024.

\bibitem[Wan et~al.(2025)Wan, Wang, Ai, Wen, Mao, Xie, Chen, Yu, Zhao, Yang, et~al.]{wan2.1}
Team Wan, Ang Wang, Baole Ai, Bin Wen, Chaojie Mao, Chen-Wei Xie, Di Chen, Feiwu Yu, Haiming Zhao, Jianxiao Yang, et~al.
\newblock Wan: {O}pen and {A}dvanced {L}arge-{S}cale {V}ideo {G}enerative {M}odels.
\newblock \emph{arXiv preprint arXiv:2503.20314}, 2025.

\bibitem[Wang et~al.(2025{\natexlab{a}})Wang, Ai, Wen, Mao, Xie, Chen, Yu, Zhao, Yang, Zeng, et~al.]{wang2025wan}
Ang Wang, Baole Ai, Bin Wen, Chaojie Mao, Chen-Wei Xie, Di Chen, Feiwu Yu, Haiming Zhao, Jianxiao Yang, Jianyuan Zeng, et~al.
\newblock Wan: Open and advanced large-scale video generative models.
\newblock \emph{CoRR}, 2025{\natexlab{a}}.

\bibitem[Wang et~al.(2025{\natexlab{b}})Wang, Cheng, Zhang, Han, and Gui]{lota2025}
Hongsong Wang, Renxi Cheng, Yang Zhang, Chaolei Han, and Jie Gui.
\newblock Lota: Bit-planes guided ai-generated image detection.
\newblock In \emph{Proceedings of the IEEE/CVF International Conference on Computer Vision}, pages 17246--17255, 2025{\natexlab{b}}.

\bibitem[Wang et~al.(2023{\natexlab{a}})Wang, Yuan, Chen, Zhang, Wang, and Zhang]{modelscope}
Jiuniu Wang, Hangjie Yuan, Dayou Chen, Yingya Zhang, Xiang Wang, and Shiwei Zhang.
\newblock Modelscope text-to-video technical report.
\newblock \emph{arXiv preprint arXiv:2308.06571}, 2023{\natexlab{a}}.

\bibitem[Wang et~al.(2023{\natexlab{b}})Wang, Yang, Tuo, He, Zhu, Fu, and Liu]{hd_vg_130m}
Wenjing Wang, Huan Yang, Zixi Tuo, Huiguo He, Junchen Zhu, Jianlong Fu, and Jiaying Liu.
\newblock Videofactory: Swap attention in spatiotemporal diffusions for text-to-video generation.
\newblock \emph{arXiv preprint arXiv:2305.10874}, 2023{\natexlab{b}}.

\bibitem[Wang et~al.(2023{\natexlab{c}})Wang, Bao, Zhou, Wang, Hu, Chen, and Li]{wang2023dire}
Zhendong Wang, Jianmin Bao, Wengang Zhou, Weilun Wang, Hezhen Hu, Hong Chen, and Houqiang Li.
\newblock Dire for diffusion-generated image detection.
\newblock In \emph{IEEE/CVF Conference on Computer Vision and Pattern Recognition}, pages 22445--22455, 2023{\natexlab{c}}.

\bibitem[Wu et~al.(2023)Wu, Ge, Wang, Lei, Gu, Shi, Hsu, Shan, Qie, and Shou]{wu2023tune}
Jay~Zhangjie Wu, Yixiao Ge, Xintao Wang, Stan~Weixian Lei, Yuchao Gu, Yufei Shi, Wynne Hsu, Ying Shan, Xiaohu Qie, and Mike~Zheng Shou.
\newblock Tune-a-video: One-shot tuning of image diffusion models for text-to-video generation.
\newblock In \emph{Proceedings of the IEEE/CVF International Conference on Computer Vision}, pages 7623--7633, 2023.

\bibitem[Xia et~al.(2024)Xia, Chen, Wu, Li, Hung, Zhan, He, and Zhou]{musev}
Zhiqiang Xia, Zhaokang Chen, Bin Wu, Chao Li, Kwok-Wai Hung, Chao Zhan, Yingjie He, and Wenjiang Zhou.
\newblock Musev: Infinite-length and high fidelity virtual human video generation with visual conditioned parallel denoising.
\newblock \emph{arxiv}, 2024.

\bibitem[Xing et~al.(2024{\natexlab{a}})Xing, Xia, Zhang, Chen, Yu, Liu, Liu, Wang, Shan, and Wong]{xing2023dynamicrafter}
Jinbo Xing, Menghan Xia, Yong Zhang, Haoxin Chen, Wangbo Yu, Hanyuan Liu, Gongye Liu, Xintao Wang, Ying Shan, and Tien-Tsin Wong.
\newblock Dynamicrafter: Animating open-domain images with video diffusion priors.
\newblock In \emph{European Conference on Computer Vision}, pages 399--417. Springer, 2024{\natexlab{a}}.

\bibitem[Xing et~al.(2024{\natexlab{b}})Xing, Xia, Zhang, Chen, Yu, Liu, Liu, Wang, Shan, and Wong]{xing2024dynamicrafter}
Jinbo Xing, Menghan Xia, Yong Zhang, Haoxin Chen, Wangbo Yu, Hanyuan Liu, Gongye Liu, Xintao Wang, Ying Shan, and Tien-Tsin Wong.
\newblock Dynamicrafter: Animating open-domain images with video diffusion priors.
\newblock In \emph{European Conference on Computer Vision}, pages 399--417. Springer, 2024{\natexlab{b}}.

\bibitem[Yan et~al.(2024{\natexlab{a}})Yan, Li, Cai, Hao, Jiang, Hu, and Xie]{yan2024sanity}
Shilin Yan, Ouxiang Li, Jiayin Cai, Yanbin Hao, Xiaolong Jiang, Yao Hu, and Weidi Xie.
\newblock A sanity check for ai-generated image detection.
\newblock \emph{arXiv preprint arXiv:2406.19435}, 2024{\natexlab{a}}.

\bibitem[Yan et~al.(2024{\natexlab{b}})Yan, Zhao, Chen, Guo, Fu, Yao, Ding, and Yuan]{yan2024generalizing}
Zhiyuan Yan, Yandan Zhao, Shen Chen, Mingyi Guo, Xinghe Fu, Taiping Yao, Shouhong Ding, and Li Yuan.
\newblock Generalizing deepfake video detection with plug-and-play: Video-level blending and spatiotemporal adapter tuning.
\newblock \emph{arXiv preprint arXiv:2408.17065}, 2024{\natexlab{b}}.

\bibitem[Yang et~al.(2020)Yang, Xu, Shi, Dai, and Zhou]{tpn}
Ceyuan Yang, Yinghao Xu, Jianping Shi, Bo Dai, and Bolei Zhou.
\newblock Temporal pyramid network for action recognition.
\newblock In \emph{Proceedings of the IEEE Conference on Computer Vision and Pattern Recognition (CVPR)}, 2020.

\bibitem[Yang et~al.(2019)Yang, Li, and Lyu]{yang2019exposing}
Xin Yang, Yuezun Li, and Siwei Lyu.
\newblock Exposing deep fakes using inconsistent head poses.
\newblock In \emph{IEEE international conference on acoustics, speech and signal processing}, pages 8261--8265. IEEE, 2019.

\bibitem[Yang et~al.(2024)Yang, Teng, Zheng, Ding, Huang, Xu, Yang, Hong, Zhang, Feng, et~al.]{yang2024cogvideox}
Zhuoyi Yang, Jiayan Teng, Wendi Zheng, Ming Ding, Shiyu Huang, Jiazheng Xu, Yuanming Yang, Wenyi Hong, Xiaohan Zhang, Guanyu Feng, et~al.
\newblock Cogvideox: Text-to-video diffusion models with an expert transformer.
\newblock \emph{arXiv preprint arXiv:2408.06072}, 2024.

\bibitem[Yang et~al.(2025)Yang, Teng, Zheng, Ding, Huang, Xu, Yang, Hong, Zhang, Feng, et~al.]{Cog}
Zhuoyi Yang, Jiayan Teng, Wendi Zheng, Ming Ding, Shiyu Huang, Jiazheng Xu, Yuanming Yang, Wenyi Hong, Xiaohan Zhang, Guanyu Feng, et~al.
\newblock {CogVideoX: Text-to-Video Diffusion Models with An Expert Transformer}.
\newblock In \emph{International Conference on Learning Representations}, 2025.

\bibitem[Yu et~al.(2021)Yu, Li, Li, Lu, and Zhou]{yu2021frequency}
Bingyao Yu, Wanhua Li, Xiu Li, Jiwen Lu, and Jie Zhou.
\newblock Frequency-aware spatiotemporal transformers for video inpainting detection.
\newblock In \emph{Proceedings of the IEEE/CVF International Conference on Computer Vision}, pages 8188--8197, 2021.

\bibitem[Yuan et~al.(2024)Yuan, Chen, Li, Jia, He, Wang, and Sun]{mora}
Zhengqing Yuan, Ruoxi Chen, Zhaoxu Li, Haolong Jia, Lifang He, Chi Wang, and Lichao Sun.
\newblock Mora: Enabling generalist video generation via a multi-agent framework, 2024.

\bibitem[Zhang et~al.(2025)Zhang, Lian, Yang, Li, Pang, Liu, Han, Li, and Tan]{zhangphysics}
Shuhai Zhang, ZiHao Lian, Jiahao Yang, Daiyuan Li, Guoxuan Pang, Feng Liu, Bo Han, Shutao Li, and Mingkui Tan.
\newblock Physics-driven spatiotemporal modeling for ai-generated video detection.
\newblock In \emph{Annual Conference on Neural Information Processing Systems}, 2025.

\bibitem[Zhang et~al.(2019)Zhang, Karaman, and Chang]{zhang2019detecting}
Xu Zhang, Svebor Karaman, and Shih-Fu Chang.
\newblock Detecting and simulating artifacts in gan fake images.
\newblock In \emph{IEEE International Workshop on information Forensics and Security}, pages 1--6. IEEE, 2019.

\bibitem[Zhao et~al.(2023)Zhao, Wang, Hu, Chen, Liu, and Tang]{zhao2023istvt}
Cairong Zhao, Chutian Wang, Guosheng Hu, Haonan Chen, Chun Liu, and Jinhui Tang.
\newblock Istvt: interpretable spatial-temporal video transformer for deepfake detection.
\newblock \emph{IEEE Transactions on Information Forensics and Security}, 18:\penalty0 1335--1348, 2023.

\bibitem[Zhao et~al.(2017)Zhao, Mathieu, and LeCun]{zhao2017energy}
Junbo Zhao, Michael Mathieu, and Yann LeCun.
\newblock Energy-based generative adversarial networks.
\newblock In \emph{International Conference on Learning Representations}, 2017.

\bibitem[Zheng et~al.(2025)Zheng, Suo, Lin, Zhao, Yang, Liu, Yang, Wang, and Shen]{zheng2025d3}
Chende Zheng, Ruiqi Suo, Chenhao Lin, Zhengyu Zhao, Le Yang, Shuai Liu, Minghui Yang, Cong Wang, and Chao Shen.
\newblock {D3: T}raining-free ai-generated video detection using second-order features.
\newblock In \emph{Proceedings of the IEEE/CVF International Conference on Computer Vision}, pages 12852--12862, 2025.

\bibitem[Zheng et~al.(2024{\natexlab{a}})Zheng, Peng, Yang, Shen, Li, Liu, Zhou, Li, and You]{OpenSora}
Zangwei Zheng, Xiangyu Peng, Tianji Yang, Chenhui Shen, Shenggui Li, Hongxin Liu, Yukun Zhou, Tianyi Li, and Yang You.
\newblock {Open-Sora: Democratizing Efficient Video Production for All}.
\newblock \emph{arXiv preprint arXiv:2412.20404}, 2024{\natexlab{a}}.

\bibitem[Zheng et~al.(2024{\natexlab{b}})Zheng, Peng, Yang, Shen, Li, Liu, Zhou, Li, and You]{zheng2024open}
Zangwei Zheng, Xiangyu Peng, Tianji Yang, Chenhui Shen, Shenggui Li, Hongxin Liu, Yukun Zhou, Tianyi Li, and Yang You.
\newblock Open-sora: Democratizing efficient video production for all.
\newblock \emph{arXiv preprint arXiv:2412.20404}, 2024{\natexlab{b}}.

\bibitem[Zhou et~al.(2018)Zhou, Andonian, Oliva, and Torralba]{trn}
Bolei Zhou, Alex Andonian, Aude Oliva, and Antonio Torralba.
\newblock Temporal relational reasoning in videos.
\newblock \emph{European Conference on Computer Vision}, 2018.

\bibitem[Zi et~al.(2020)Zi, Chang, Chen, Ma, and Jiang]{zi2020wilddeepfake}
Bojia Zi, Minghao Chang, Jingjing Chen, Xingjun Ma, and Yu-Gang Jiang.
\newblock Wilddeepfake: A challenging real-world dataset for deepfake detection.
\newblock In \emph{Proceedings of the ACM International Conference on Multimedia}, pages 2382--2390, 2020.

\end{thebibliography}
}


\end{document}